\documentclass{article}

\usepackage[final]{neurips_robustseq_2022}

\usepackage{tikz}
\usepackage{booktabs}
\usepackage{subcaption}
 
\usepackage[utf8]{inputenc} 
\usepackage[T1]{fontenc}    
\usepackage{hyperref}       
\usepackage{url}            
\usepackage{booktabs}       
\usepackage{amsfonts}       
\usepackage{nicefrac}       
\usepackage{microtype}      
\usepackage{xcolor}         
\usepackage{amsmath}

\title{Are Deep Sequence Classifiers Good at Non-Trivial Generalization?}

\author{Francesco Cazzaro \\ Universitat Politècnica\\ de Catalunya, \\Barcelona, Spain \\ \texttt{\{francesco.cazzaro,}
\And Ariadna Quattoni  \\ Universitat Politècnica\\ de Catalunya, \\Barcelona, Spain \\ \texttt{ariadna.julieta.quattoni\}@upc.edu }  \And Xavier Carreras
        \\ dMetrics, \\ Brooklyn, NY
        }

\begin{document}

\maketitle

\begin{abstract}
  Recent advances in deep learning models for sequence classification have greatly improved their classification accuracy, specially when large training sets are available. However, several works have suggested that under some settings the predictions made by these models are poorly calibrated. In this work we study binary sequence classification problems and we look at model calibration from a different perspective by asking the question: Are deep learning models capable of learning the underlying  target class distribution? We focus on sparse sequence classification, that is problems in which the target class is rare and compare three deep learning sequence classification models.  We develop an evaluation that measures how well a classifier is learning the target class distribution. In addition, our evaluation disentangles good performance achieved by mere compression of the training sequences versus performance achieved by proper model generalization. Our results suggest that in this binary setting the deep-learning models are indeed able to learn the underlying class distribution in a non-trivial manner, i.e. by proper generalization beyond data compression.
\end{abstract}

\section{Introduction}\label{sec:intro}



Recent advances in deep learning models for sequence classification have greatly improved their classification accuracy, specially when large training sets are available. However, several works have suggested that under some settings the predictions made by these models are poorly calibrated: a model might provide in general correct predictions but give bad estimates of the confidence scores $\Pr(y|x)$. Having well calibrated estimates of $\Pr(y|x)$ is crucial to many machine learning applications. For example, this is critical in systems in which an end user needs to make a decision based on the prediction of the model. Here a good estimate is as important as a good prediction.


In this work we study binary sequence classification with a focus on sparse sequence classification tasks in which only a very small fraction of the sequences in the domain belong to the target class. These types of problems naturally arise in several applications, for example in NLP or in Computational Biology where distinguishing relevant segments in DNA sequences is challenging as it is postulated that only about \mbox{1--3\%} of segments have any biological significance \citep{eskin2002sparse}.

While traditional evaluations on model calibration have focused on the quality of the conditional prediction $\Pr_M(y|x)$ in this paper we look at model evaluation from a different perspective by asking the question: Are deep learning models capables of learning the underlying target distribution? The main idea is quite simple, if we have access to large quantities of unlabeled data that accurately represent the input distribution we can estimate the joint distribution $\Pr_M(y,x)$ implicitly induced by the conditional model and then we can compare it to the true distribution. 


We conduct experiments using three deep-learning models: an RNN, a Transformer and a WFA (i.e. Weighted Finite Automata, a form of RNN with linear activation). Our results suggest that in the binary classification setting the deep-learning models are indeed able to learn the underlying class distribution in a non-trivial manner. These results seem to be consistent with previous studies showing that deep-learning models provide calibrated predictions for binary classification problems \citep{10.1145/1102351.1102430}. Our main contributions are:
1) We present a novel evaluation framework for sequence prediction models. By exploiting unlabeled data, we evaluate the model with respect to the implicitly induced joint distribution.
2) Our evaluation distinguishes performance over components of the distribution seen in training data as well as unseen components explicitly differentiating compression from generalization.
3) Our experiments on a sparse sequence classification task show that deep learning architectures are able to induce good distributions in a non-trivial manner.


\section{Related Work}

In the recent literature several works have addressed the problem of model calibration. Some of these studies have shown that for some deep learning architectures, the predictions produced by the models are not well calibrated \citep{10.5555/3305381.3305518} \citep{DBLP:journals/corr/abs-1903-00802} especially for non pre-trained transformers \citep{desai-durrett-2020-calibration}. In contrast, for the case of  binary classification some previous work has suggested that they are indeed well calibrated \citep{10.1145/1102351.1102430}. 
To our knowledge we are the first ones to study calibration by looking at the joint distribution $\Pr(x,y)$ induced by the learned classifier instead of the quality of the conditional class predictions.

Calibration aside, several works have compared the classification performance of different deep-learning architectures: CNNs and RNNs \citep{DBLP:journals/corr/JozefowiczVSSW16} \citep{DBLP:journals/corr/0001KYS17}, Transformers and RNNs \citep{9003750} \citep{lakew-etal-2018-comparison}, Transformers and CNNs \citep{50650} \citep{pinto2021are} \citep{DBLP:journals/corr/abs-2111-05464}, WFA to CNNs \citep{quattoni-carreras-2020-comparison} and WFA to RNNs \citep{quattoni-carreras-2019-interpolated}.

Finally, an orthogonal related problem is that of developing deep learning models for density estimation. These include Transformers \citep{DBLP:journals/corr/abs-2004-02441}, Autoregressive Networks \citep{10.5555/2946645.3053487} \citep{oliva2018transformation} and Flow Models \citep{NEURIPS2019_7ac71d43} \citep{pmlr-v115-de-cao20a}. 

\section{Evaluating Deep-Learning Model as Moment Predictors}

Our goal is to evaluate models defined over sequences of discrete symbols. More precisely we consider an alphabet $\Sigma$ and the set of all possible sequences $\Sigma^\star$. In general, we can think of probabilistic binary sequence classifiers as functions from $\Sigma^{\star} \to [0,1]$. In our setting we assume that we have access to a large set of sequences $U=\{x^{(1)}, \ldots, x^{(u)}\}$ sampled according to the underlying distribution over $\Sigma^{\star}$. We can think of $U$ as a large set of unlabeled sequences that represent the domain of the sequence classification task. The target class $Y$ that we wish to learn is a subset of sequences in $U$, that is $Y \subset U$. We are particularly interested in cases where the target class is rare, that is cases in which $|Y|$ is significantly smaller than $|U|$. 

We will create a labeled training set $T=\{(x^{(1)},y^{(1)}), \ldots , (x^{(m)},y^{(m)})\}$ of size $m$ by sampling sequences from $U$ and labeling $y=1$ as positive instances if they appear in $Y$, and otherwise as negative instances $y=0$. Section 4 contains further details about how we create the task data from an existing dataset, for various training sizes $m$. Given a training set we will train a sequence classifier $M$ that defines a distribution $\Pr_{M}(y|x)$.

In this work, we consider three sequence classification models: an \textbf{RNN}, a \textbf{Transformer} and a \textbf{WFA}. As recurrent neural network (RNN) we employ a multi-layer LSTM \citep{10.1162/neco.1997.9.8.1735} with a binary classification feed-forward layer on top. For the Transformer we select the BERT \citep{devlin-etal-2019-bert} architecture, we don’t use the pre-trained weights and we expand the embeddings with new randomly initialized ones to deal with the protein dataset vocabulary. Both models output the conditional probability $\Pr(l|x)$ where $x$ is a sentence and $l$ a label. We also evaluate a WFA (Weighted Finite Automata; which in essence is an RNN with linear activation functions \citep{pmlr-v89-rabusseau19a}) employing the ensemble proposed in \citet{quattoni-carreras-2020-comparison}.

\subsection{Evaluation Metrics}
\label{sec:evaluation_metrics}
The sequence model $M$ that we wish to evaluate defines a conditional distribution $\Pr_{M}(y|x)$. Generally, we would like to compute the true error of $M$ by evaluating the model on the true distribution $\Pr(x, y)$. In this section, we will use the set $U$ as a proxy for $\Pr(x)$, and specifically we will estimate the \emph{moments} of $M$ on the joint distribution $\Pr_{M}(x, y)$ that is implicitly induced by the model over the target class.\footnote{We highlight that the set $U$ is assumed to be sampled from the true $\Pr(x)$ and will contain repetitions of each distinct $x$ proportional to its likelihood. And so does $Y$. This detail is relevant because we can think of $U$ as an empirical estimate of $\Pr(x)$, which is necessary for the definitions of moments in this section. }
More precisely, consider the set: 
\begin{equation}
Z_{n} =\{z_{n} \mid z_n \in \Sigma^n , \exists x \in U \; \text{such\ that} \; z_{n} \in x\}
\end{equation}
where $z_{n}$ is a sub-sequence of size $n$, $z_n \in x$ indicates that $z_n$ is a sub-sequence of a domain sequence $x$, and therefore $Z_{n}$ is the set of all sub-sequences of size $n$ observed in $U$. 
We use these sub-sequences as the support set to compute moments of a model, and we define the \emph{moment} function $E_M: z_{n} \times U \to \mathbb{R}$ as:
\begin{equation}
    E_{M}(z_{n}, U)={\frac{1}{|U|}\sum_{x\in U} \Pr\!\!\,_{M}(y|x) \ \#[z_{n} \in x]}
\end{equation}
where $z_{n}$ is a sub-sequence of length $n$ and $\#[z_{n} \in x]$ is a function that counts the number of times that $z_{n}$ appears in sequence $x$. From now on, we will assume a fixed $U$ and drop it from the expression, and we will refer to the sub-sequences $z_n$ and their values $E_M(z_{n})$ as the \emph{moments} of $M$. 
Since we assume that $U$ is a good representation of the the domain, $E_{M}(z_{n})$ can be regarded as an estimate of the expected number of times that we should observe $z_{n}$ in a sample from $\Pr_M(x,y)$. 

Given sets $U$ and $Y$, the \emph{moments} from the true joint distribution $\Pr_Y(x,y)$ can be computed similarly as:
\begin{equation}
    E_{Y}(z_{n}) = \frac{1}{|U|} \sum_{x\in Y}{\#[z_{n} \in x]}
\end{equation}

Putting it all together, if we have a domain represented by a set of unlabeled sequences $U$, a subset of target sequences $Y$ and a model $\Pr_M(y|x)$, we can compute the model's \emph{moment} function $E_{M}(z_{n})$ and compare it to the true \emph{moments} $E_{Y}(x_{n})$. 
Notice that for a model that learns the target class perfectly (i.e. a model that gives accurate predictions and that it is perfectly calibrated) we will have $E_{M}(z_{n})=E_{Y}(z_{n})$ for all \emph{moments} in $N$.

Now that we have the necessary functions, we propose to evaluate a model $M$ by comparing $E_{M}$ and $E_{Y}$. To compare the \emph{moment} functions we propose the following metrics:

\newcommand{\ranks}{\mathrm{rank}}

     \textbf{MSPC}: This metric measures the Spearman rank correlation between gold and model \emph{moments} of a fixed length. The Spearman correlation coefficient measures the strength and direction of association between ranked variables. In our evaluation a high MSPC means that the model sorts the \emph{moments} of the distribution in a way that is similar to the gold ordering. The MSPC is computed as:
    \begin{equation}
        \frac{\mathrm{cov}(\ranks[E_{M}(z_n)],\ranks[E_{Y}(z_n)])}{\sigma(\ranks[E_{M}(z_n)]) \cdot \sigma(\ranks[E_{Y}(z_n)])}
    \end{equation}
    where $\ranks[E(z_n)]$ are the raw function scores for all $z_n\in Z_n$ converted to ranks.
    In essence, this metric measures the agreement on partial orderings induced by the model and gold \emph{moment} functions. That is, given any two pairs of \emph{moments} of a fixed length $z_{n}$ and $z'_{n}$ whenever $E_{Y}(z_{n})>E_{Y}(z'_{n})$ we want $E_{M}(z_{n})>E_{M}(z'_{n})$.
    
     \textbf{MSPCP}: While MSPC is a useful metric when evaluating the class distribution induced by a model, we need to keep in mind that for longer \emph{moments} $E_{Y}(z_{n})$ will be sparse and for most inputs it will be 0 since we are targeting a rare class. 
     So in this context we compute the Spearman rank correlation between gold non-zero \emph{moments} and the model prediction for those \emph{moments}, this will give an idea of how well the model would sort the \emph{moments} of the target distribution if it knew the true support of the target \emph{moments} function. 
    
     \textbf{MR}: as MSPC might be overly harsh, a complementary alternative metric in this case is the mean-rank given to the target non-zero \emph{moments}:
    \begin{equation}
     1 - \sum_{z_n\ :\ E_{Y}(z_n)>0}\frac{\ranks[E_{M}(z_n)]}{|Z_n|}
    \end{equation}
    This metric captures how well a model performs in the task of making the binary prediction of which \emph{moments} should be non-zero for the target class.

We believe that it is important to make a distinction between the ability of a model to compress data versus its ability to generalize to unseen data. A natural way to achieve this goal is to make a distinction between \emph{moments} that have been seen in the training set, versus \emph{moments} that have not been seen. That is, for any training set $T$ and \emph{moment} length $n$ we can define a subset 
$Z_{n}^{T} \subseteq Z_{n}$: 
\begin{equation}
    Z_{n}^{T}=\{ z_n \in Z_{n} : z_n \; \text{appears\ in}\; T\}
\end{equation}
Given training set $T$ we can define its analogous \emph{unseen} evaluation by computing the metric over $Z_{n} \setminus Z_{n}^{T}$. We call these new metrics: \mbox{\textbf{MSPC-U}}, \mbox{\textbf{MSPCP-U}} and
\mbox{\textbf{MR-U}}.

In order to evaluate the generalization ability of a model we will also consider a baseline model that memorizes the training data and makes predictions with the observed training \emph{moments}. More precisely, given a training set $T$ the baseline will compute the \emph{moments} function as:
\begin{equation}
    E_{B_{T}}(z_{n})=\frac{1}{|U|} \sum_{(x,y) \in T :y=1}{\#[z_{n} \in x]}
\end{equation}

\section{Experiments}

To generate the data used in our experiments we use the PFAM dataset \citep{10.1093/nar/gky995}. This dataset consists of aminoacid sequences pertaining to different protein fami\-lies. Deep-models have been shown to be very successful at classifying the complete sequences \citep{Bileschi626507} \citep{szalkai2018near}.  To derive a more challenging task we decide to work with segments and we generate the unlabeled set $U$ by sampling sequence segments of size 5. 

To create our target class $Y$ we pick a protein-family and select all segments that appear in sequences from that family. We select as our target the most rare or sparse family based on the ratio $|Y|/|U|$. We also select a second family as validation data. We create $5$ labeled training sets $T=\{(x^{(1)},y^{(1)}), \ldots , (x^{(m)},y^{(m)})\}$ by sampling sequences with $m \in \{ 400, 1000, 2000, 4000, 8000\}$. For the optimization details refer to appendix \ref{sec:optimization_details}.

\subsection{Results}

\begin{figure*}[!h]
\centering
\begin{subfigure}[b]{0.48\textwidth}
    \centering
    \includegraphics[width=0.7\linewidth]{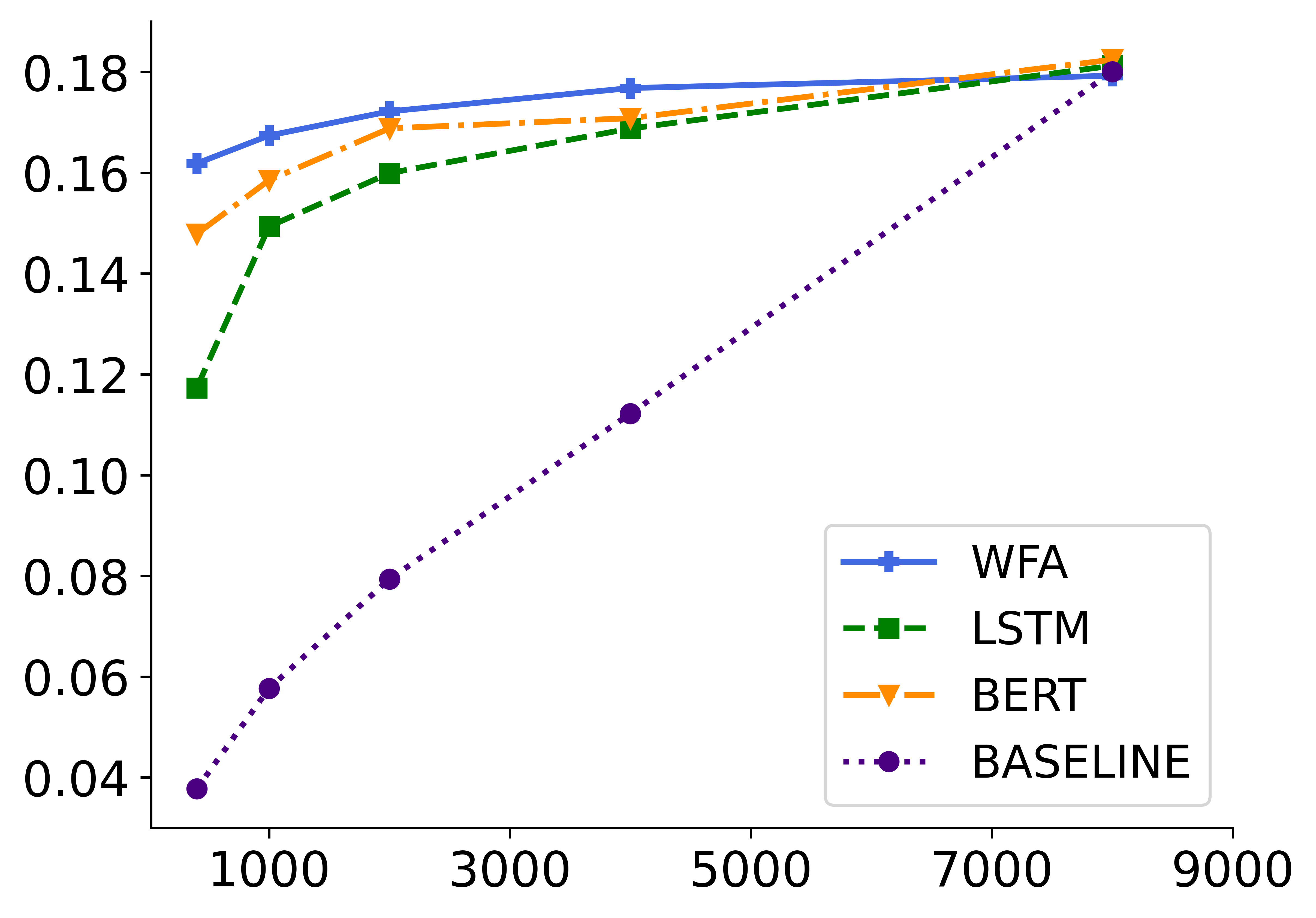} 
    \caption{MICRO MSPC}
    \label{fig:spc}
  \end{subfigure}
  \vspace{4mm}
  \begin{subfigure}[b]{0.48\textwidth}
    \centering
      \includegraphics[width=0.7\linewidth]{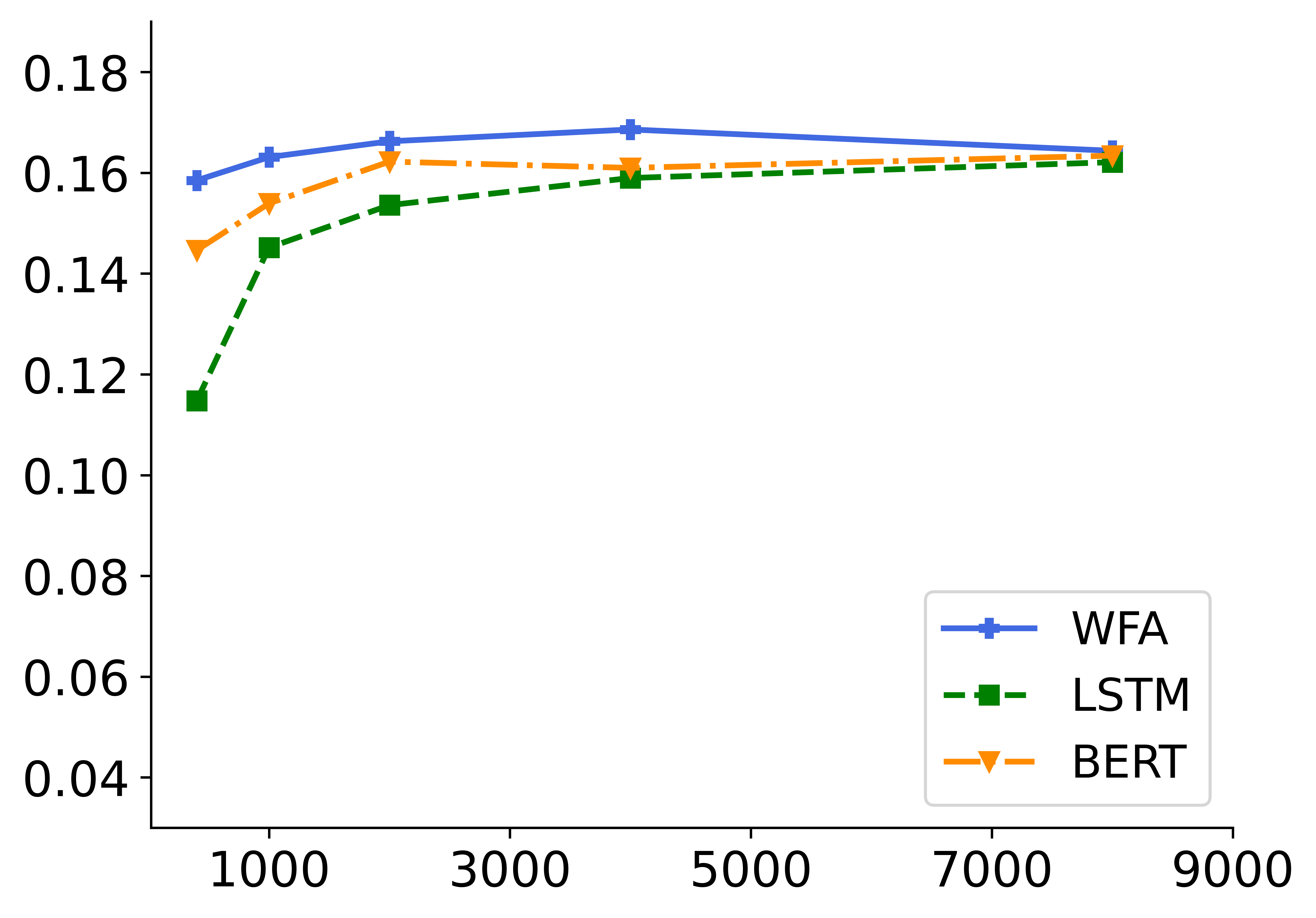} 
    \caption{MICRO MSPC-U}
  \label{fig:spc_u}
  \end{subfigure}
  \vspace{3mm}
    \begin{subfigure}[b]{0.48\textwidth}
    \centering
    \includegraphics[width=0.7\linewidth]{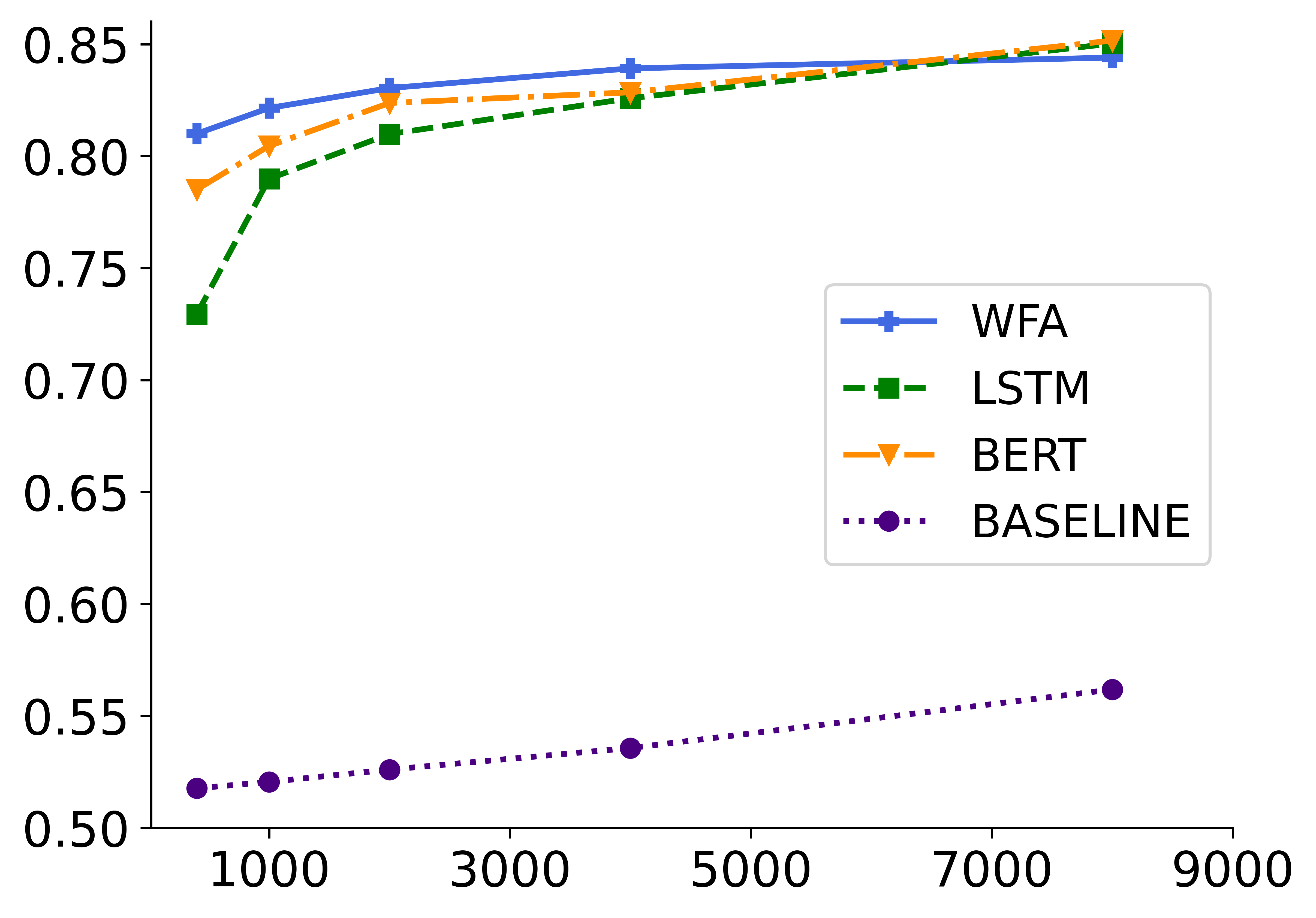} 
    \caption{MICRO MR}
    \label{fig:mr}
  \end{subfigure}
  \vspace{4mm}
  \begin{subfigure}[b]{0.48\textwidth}
    \centering
      \includegraphics[width=0.7\linewidth]{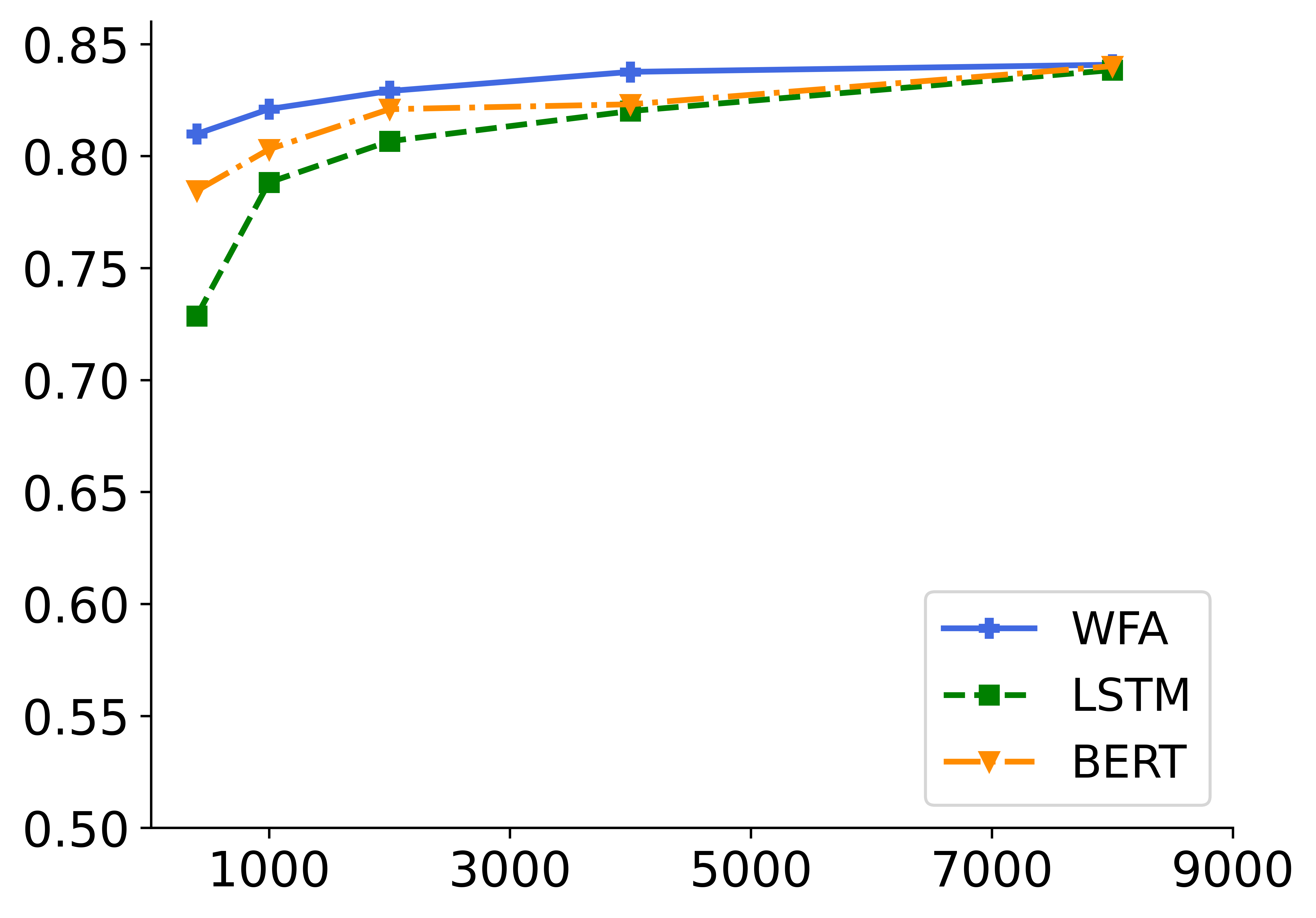} 
    \caption{MICRO MR-U}
  \label{fig:mr_u}
  \end{subfigure}
  \begin{subfigure}[b]{0.48\textwidth}
    \centering
    \includegraphics[width=0.7\linewidth]{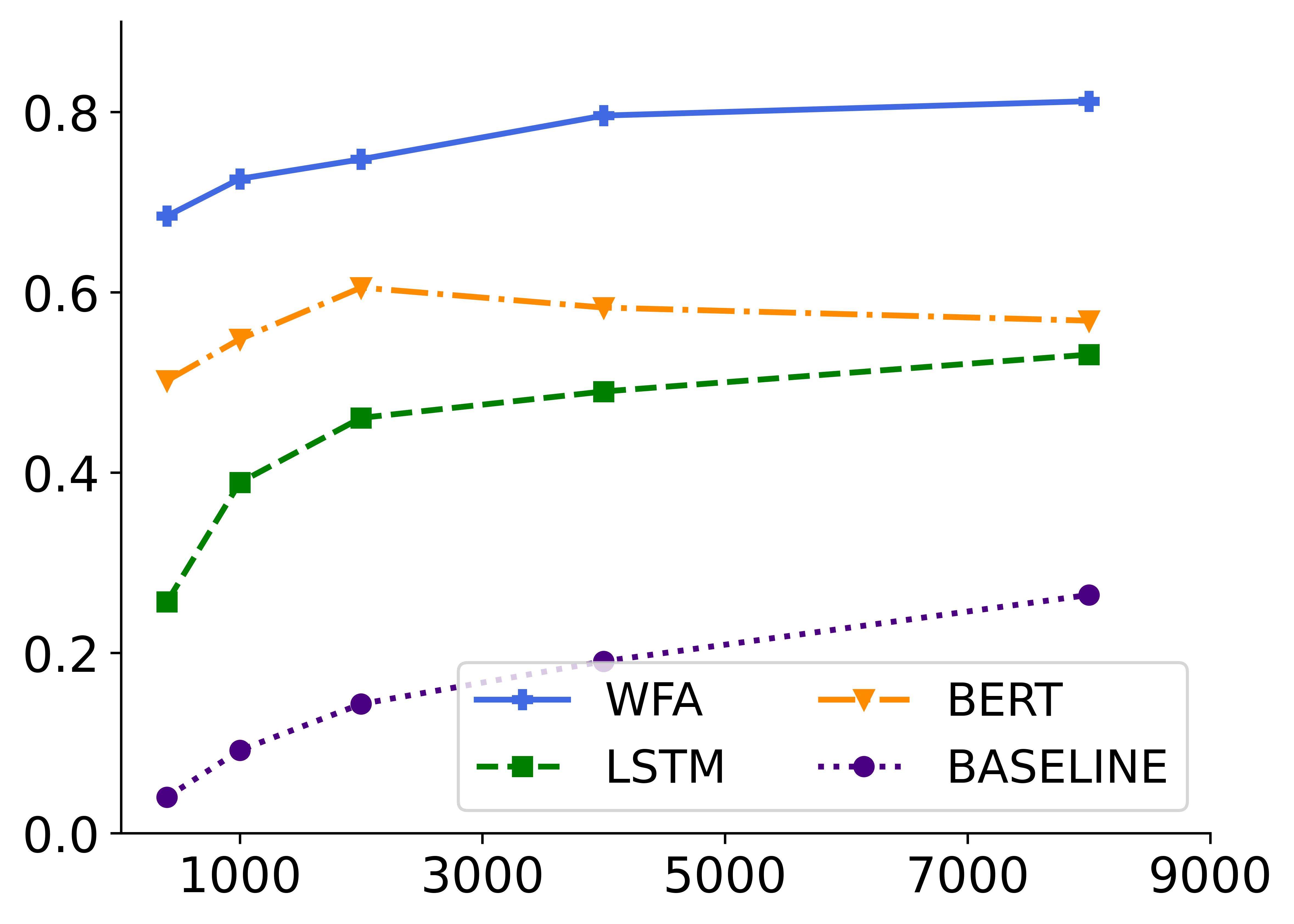} 
    \caption{MICRO MSPCP}
    \label{fig:spc_pos}
  \end{subfigure}
  \vspace{4mm}
  \begin{subfigure}[b]{0.48\textwidth}
    \centering
      \includegraphics[width=0.7\linewidth]{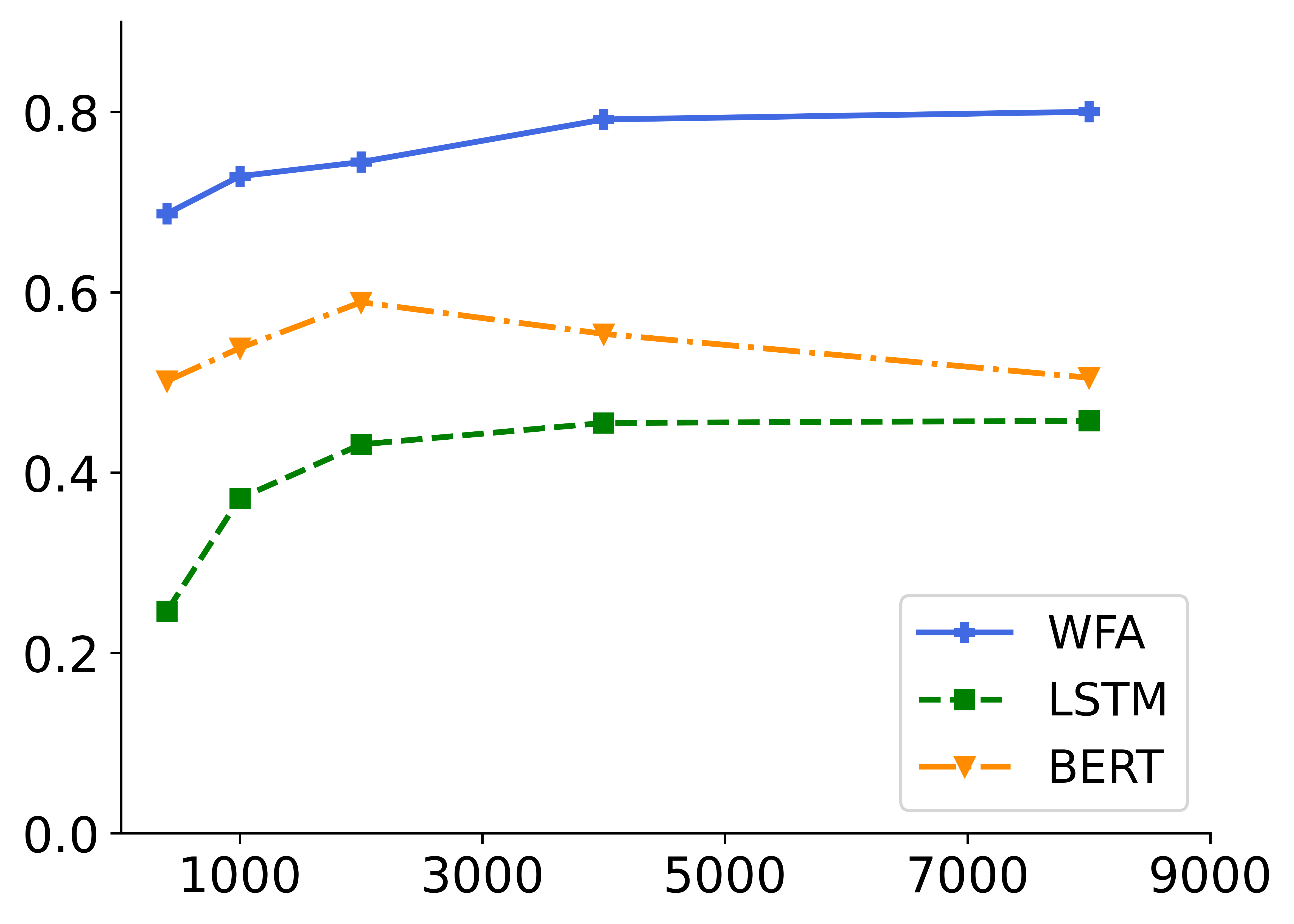} 
    \caption{MICRO MSPCP-U}
  \label{fig:spc_pos_u}
  \end{subfigure}
  \caption{MICRO MSPC, MR and MSPCP results}
\end{figure*}

To summarize the results, for each training set size and model, we compute an aggregate metric, which we refer as MICRO (MC). The MICRO score is defined as a weighted average of the scores for \emph{moments} of different lengths, where the weight is proportional to the total number of test-moments of that size. The full breakdown of results is available in appendix \ref{sec:full_tables}.

Figure \ref{fig:spc} shows MICRO MSPC for the three models and the baseline. The first observation is that all models perform significantly better than the baseline. Recall that the baseline is essentially remembering the \emph{moments} that it has observed at training, therefore it is natural that as we increase the number of training sequences its performance approaches that of the learned models. Figure \ref{fig:spc_u} shows the analogous results when MSPC is computed over unseen \emph{moments}, the baseline in this case is undefined since by definition the expectation over all unseen \emph{moments} will be zero. Still the models seem to be learning to rank unseen \emph{moments}, and therefore they are making proper generalizations.

However, one obvious observation is that the MSPC is not very high in absolute terms (appendix \ref{sec:full_tables} tables, both in the all and unseen scenarios). Looking at the absolute value of the MSPC might be misleading because we are modeling a sparse sequence classification task. In other words, the gold expectation for most of the \emph{moments} in the domain is zero and in this case the MSPC metric might be too harsh. Thus, as described in Section \ref{sec:evaluation_metrics}, we turn our attention to MR.

The MR metric reflects the model's ability to predict the non-zero \emph{moments} of the distribution and as shown in Figure \ref{fig:mr} all models learn to do this relatively well. This is also true for the generalization setting (Figure \ref{fig:mr_u}) in which models are forced to make predictions over unseen \emph{moments}.

To gain more insight into the distributions induced by the models, another way of breaking down the MSPC metric is to focus on how well a model ranks the non-zero \emph{moments} of the gold distribution. That is, in this case, we tell the model which \emph{moments} are non-zero and let the model rank them. Figure \ref{fig:spc_pos} shows MSPCP. We can see that all models learn to rank the \emph{moments} of the target class and this also holds true for unseen \emph{moments} (Figure \ref{fig:spc_pos_u}). In this evaluation scenario, the WFA seems to be better than the other two models. This is most likely due to the difference between the loss function minimized by the WFA and the loss functions minimized by the LSTM and BERT models. The WFA classifier is defined as an ensemble of class specific models, and therefore it is explicitly modeling the class distributions.

Finally we also explored correlations between different models by computing pairwise Spearman correlation between the model's predictions. We found out that the correlation is high for shorter \emph{moments} and that it decreases significantly with longer ones. Regarding pairwise associations, LSTM and BERT show a higher level of agreement, while the WFA is more correlated to BERT overall.

\section{Conclusion}

In this paper we presented a novel evaluation framework for comparing sequence classification models and for testing the extent to which they can implicitly learn the target distribution. Our results comparing three different deep-learning models suggest that the models have the ability to learn the underlying distribution. They also suggest that despite significant differences in the architectures and even in the loss functions (i.e. for the WFA), their performance is comparable.  The goal of this paper was to develop a simple way to gain insights into what is actually been learned by a given sequence classifier. But the evaluation approach itself also suggests ways in which we could exploit unlabeled data to obtain better calibrated sequence classifiers, future work should explore this possibility.

\begin{ack}
The authors gratefully acknowledge the computer resources at ARTEMISA, funded by the European Union ERDF and Comunitat Valenciana as well as the technical support provided by the Instituto de Física Corpuscular, IFIC (CSIC-UV). This work is supported by the European Research Council (ERC) under the European Union’s Horizon 2020 research and innovation program (grant agreement No.853459 ).
\end{ack}

\bibliography{myref}

\appendix

\section{Optimization Details}
\label{sec:optimization_details}
For each of our training sets $T^m=\{(x^{(1)},y^{(1)}), \ldots , (x^{(m)},y^{(m)})\}$ we train a new model and validate the parameters independently. We use a $2$ layer LSTM and we validate all parameters such as the dimension of the hidden state, of the embeddings and the learning rate with a grid search evaluated on the validation data. We proceed analogously for the BERT parameters, note that we use the \textit{bert-base} model. 

We use Adam \citep{articleadam} as optimizer employing the standard cross entropy loss function. For both LSTM and BERT we train $5$ models with different starting random seeds for each training set and we report the average of the results. 
 
\begin{table}[h!]
    \centering
    \caption{Training time and minimum number of parameters to fully compress 8000 sequences.}
    \label{tab:models_size_speed}
    \begin{tabular}{lll}
    \toprule
& \bfseries Training time (s) & \bfseries \# of parameters  \\ 
\midrule 
WFA & $67$ & $1.15 \times 10^5$\\ 
LSTM & $63$ &  $1 \times 10^6$\\ 
BERT & $305$ & $1.1 \times 10^8$ \\ 
\bottomrule 
\end{tabular}
\end{table}
 
For the WFA model the only parameter that we need to validate is the number of states $k$. Besides, as there are no randomly initialized parameters and the training is not stochastic, we don't need to perform multiple runs with multiple seeds. The WFA is trained using the spectral learning method. For a detailed and comprehensive explanation of the training method the reader is referred to \citet{Balle2014}. Furthermore, we employ the scalability techniques described in  \citet{pmlr-v54-quattoni17a}.

LSTM and BERT models were trained on a single GPU while the WFA, for which the bottleneck is computing a singular value decomposition of a large sparse matrix, was trained on a CPU. To give an idea of the relative size of each model, we computed the minimum number of parameters for which the models were able to achieve perfect classification on our biggest training set. We report the numbers in Table \ref{tab:models_size_speed}. 

\section{Data}

Here we presents some statistics about the data that we have generated. There are a total of $23$ amino acid symbols.

\begin{table}[h!]
    \centering
    \caption{Dataset statistics.}
    \label{tab:dataset_stats}
    \begin{tabular}{lll}
    \toprule
& \bfseries $\#$ of sequences & \bfseries \% wrt $|\cup_n{Z_{n}}|$ \\
    \midrule
$|\cup_n{ Z_{n}}|$ & $1.8 \times 10^6$ & - \\ 
$|Z_{1}^{Y}|$ & $2 \times 10^1$   & $86.96 \%$\\ 
$|Z_{2}^{Y}|$ & $3.9 \times 10^2$ & $86.70 \%$ \\ 
$|Z_{3}^{Y}|$ & $4.1 \times 10^3$ & $49.89 \%$ \\
$|Z_{4}^{Y}|$ & $1.5 \times 10^4$  & $10.00 \%$ \\
$|Z_{5}^{Y}|$ & $2.8 \times 10^4$ & $1.68 \%$ \\
\bottomrule
\end{tabular}
\end{table}

\section{Models correlation}
Here we show a figure displaying the pairwise Spearman correlation between the model’s predictions. $N_i$ stands for the different ngram lengths.

\begin{figure}[]
  \centering
  \includegraphics[width=0.60\linewidth]{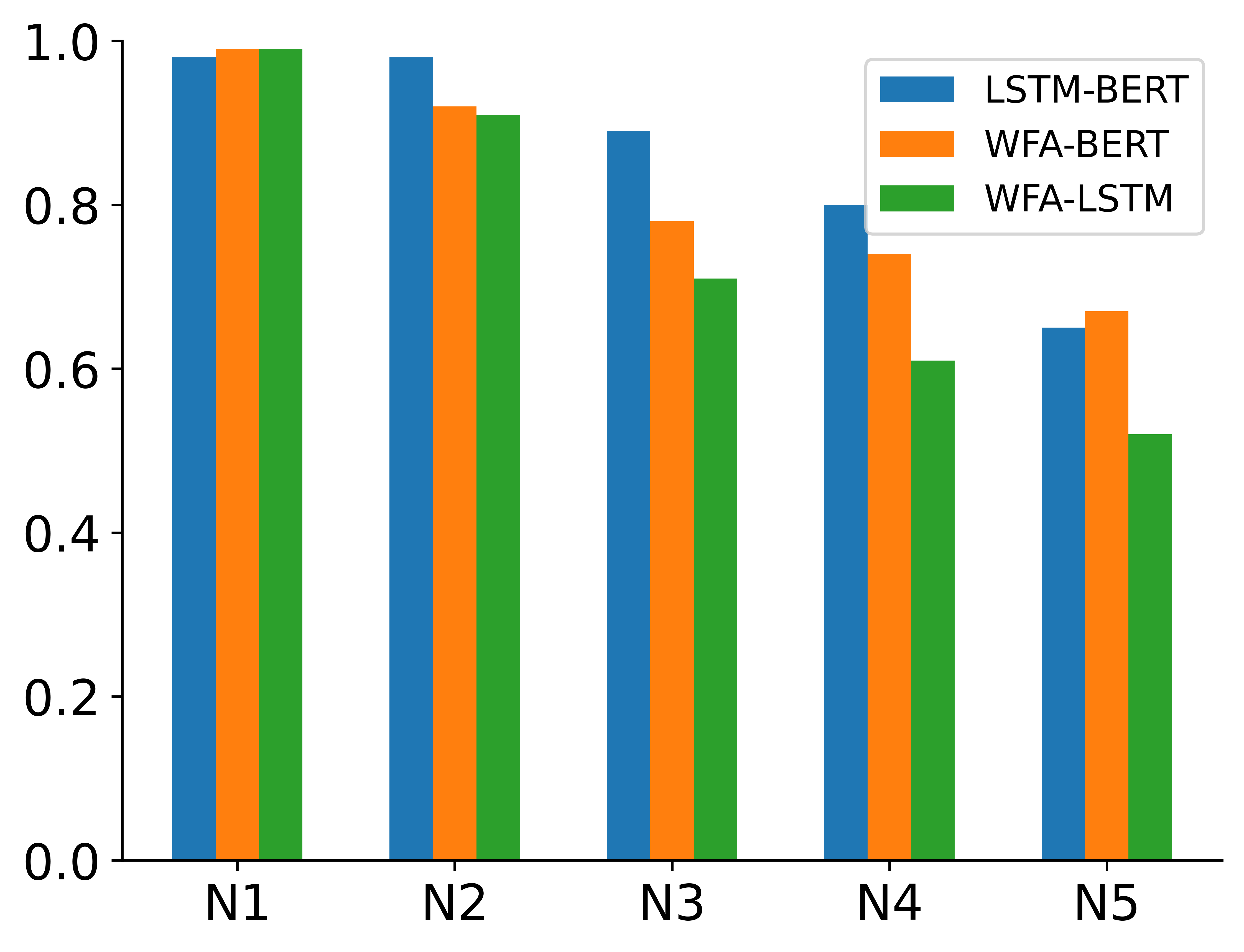}
  \caption{MSPC between models.}\label{fig:models_correlation}
\end{figure}

\vspace{46pt}

\section{Result Tables}
\label{sec:full_tables}

\begin{table}[h]
    \centering
    \caption{MSPC, 400 Training Sequences.}
    \label{tab:spc_all_410}
    \begin{tabular}{lllll}
    \toprule
    & \bfseries WFA & \bfseries LSTM & \bfseries BERT & \bfseries BASELINE \\ 
    \midrule 
    N1 & 0.92878 & 0.87715 & 0.88803 & 0.88812 \\ 
    N2 & 0.86161 & 0.7269 & 0.81566 & 0.62104 \\ 
    N3 & 0.64555 & 0.41444 & 0.57328 & 0.16744 \\ 
    N4 & 0.34245 & 0.22023 & 0.30406 & 0.051655 \\ 
    N5 & 0.14245 & 0.10607 & 0.13102 & 0.03562 \\ 
    MICRO & 0.16185 & 0.1173 & 0.14786 & 0.037732 \\ 
    \bottomrule 
\end{tabular}
\end{table}

\begin{table}[]
    \centering
    \caption{MSPC, 1000 Training Sequences.}
    \label{tab:spc_all_1026}
    \begin{tabular}{lllll}
    \toprule
    & \bfseries WFA & \bfseries LSTM & \bfseries BERT & \bfseries BASELINE \\ 
    \midrule
    N1 & 0.94362 & 0.93947 & 0.95351 & 0.90198 \\ 
    N2 & 0.88806 & 0.7933 & 0.86798 & 0.75504 \\ 
    N3 & 0.67164 & 0.51563 & 0.628 & 0.26579 \\ 
    N4 & 0.3531 & 0.29955 & 0.33486 & 0.086673 \\ 
    N5 & 0.14741 & 0.13336 & 0.13959 & 0.053745 \\ 
    MICRO & 0.1674 & 0.14934 & 0.15853 & 0.05768 \\ 
    \bottomrule 
    \end{tabular}
    \label{tab:spc_all_1026}
\end{table}

\begin{table}[]
    \centering
    \caption{MSPC, 2000 Training Sequences.}
    \label{tab:spc_all_2169}
    \begin{tabular}{lllll}
    \toprule
    & \bfseries WFA &  \bfseries LSTM & \bfseries BERT & \bfseries BASELINE \\ 
    \midrule 
    N1 & 0.96439 & 0.95964 & 0.94995 & 0.86733 \\ 
    N2 & 0.92286 & 0.87423 & 0.90971 & 0.75083 \\ 
    N3 & 0.70414 & 0.58983 & 0.68951 & 0.34454 \\ 
    N4 & 0.36564 & 0.3234 & 0.36066 & 0.11984 \\ 
    N5 & 0.15135 & 0.1424 & 0.14818 & 0.07408 \\ 
    MICRO & 0.1722 & 0.15994 & 0.16882 & 0.079362 \\ 
    \bottomrule 
    \end{tabular}
\end{table}

\begin{table}[]
    \centering
    \caption{MSPC, 4000 Training Sequences.}
    \label{tab:spc_all_3901}
    \begin{tabular}{lllll}
    \toprule
    & \bfseries WFA & \bfseries LSTM & \bfseries BERT & \bfseries BASELINE \\ 
    \midrule
    N1 & 0.97824 & 0.96855 & 0.96439 & 0.92673 \\ 
    N2 & 0.94558 & 0.8781 & 0.92219 & 0.79916 \\ 
    N3 & 0.73057 & 0.63392 & 0.68747 & 0.442 \\ 
    N4 & 0.37522 & 0.34892 & 0.35885 & 0.16727 \\ 
    N5 & 0.15539 & 0.14952 & 0.15057 & 0.10524 \\ 
    MICRO & 0.17682 & 0.16879 & 0.17084 & 0.1122 \\ 
    \bottomrule 
    \end{tabular}
\end{table}

\begin{table}[]
    \centering
    \caption{MSPC, 8000 Training Sequences.}
    \label{tab:spc_all_8379}
    \begin{tabular}{lllll}
    \toprule
    & \bfseries WFA & \bfseries LSTM & \bfseries BERT & \bfseries BASELINE \\ 
    \midrule 
    N1 & 0.98714 & 0.98398 & 0.98022 & 0.94554 \\ 
    N2 & 0.94925 & 0.92542 & 0.93935 & 0.82521 \\ 
    N3 & 0.73837 & 0.69595 & 0.72027 & 0.55846 \\ 
    N4 & 0.37987 & 0.37275 & 0.38072 & 0.25477 \\ 
    N5 & 0.15761 & 0.16066 & 0.16111 & 0.17104 \\ 
    MICRO & 0.17927 & 0.18125 & 0.18245 & 0.18006 \\ 
    \bottomrule 
    \end{tabular}
\end{table}

\begin{table}[]
    \centering
    \caption{MR, 400 Training Sequences.}
    \label{tab:my_label}
    \begin{tabular}{lllll}
    \toprule
    & \bfseries WFA & \bfseries LSTM & \bfseries BERT & \bfseries BASELINE \\ 
    \midrule 
    N4 & 0.78962 & 0.68841 & 0.75878 & 0.52947 \\ 
    N5 & 0.81186 & 0.73307 & 0.7875 & 0.51656 \\ 
    MICRO & 0.80996 & 0.72927 & 0.78505 & 0.51765 \\ 
    \bottomrule 
    \end{tabular}
\end{table}

\begin{table}[]
    \centering
    \caption{MR, 1000 Training Sequences.}
    \label{tab:my_label}
    \begin{tabular}{lllll}
    \toprule
    & \bfseries WFA & \bfseries LSTM & \bfseries BERT & \bfseries BASELINE \\ 
    \midrule 
    N4 & 0.79955 & 0.75561 & 0.78494 & 0.53438 \\ 
    N5 & 0.82346 & 0.793 & 0.80632 & 0.51921 \\ 
    MICRO & 0.82143 & 0.78982 & 0.8045 & 0.5205 \\ 
    \bottomrule 
    \end{tabular}
\end{table}

\begin{table}[]
    \centering
    \caption{MR, 2000 Training Sequences.}
    \label{tab:my_label}
    \begin{tabular}{lllll}
    \toprule
    & \bfseries WFA & \bfseries LSTM & \bfseries BERT & \bfseries BASELINE \\ 
    \midrule 
    N4 & 0.81061 & 0.77585 & 0.80705 & 0.54377 \\ 
    N5 & 0.83217 & 0.81294 & 0.82522 & 0.52434 \\ 
    MICRO & 0.83034 & 0.80979 & 0.82368 & 0.526 \\ 
    \bottomrule 
    \end{tabular}
\end{table}

\begin{table}[]
    \centering
    \caption{MR, 4000 Training Sequences.}
    \label{tab:my_label}
    \begin{tabular}{lllll}
    \toprule
    & \bfseries WFA & \bfseries LSTM & \bfseries BERT & \bfseries BASELINE \\ 
    \midrule 
    N4 & 0.81873 & 0.79771 & 0.80568 & 0.55984 \\ 
    N5 & 0.84102 & 0.82841 & 0.83056 & 0.53338 \\ 
    MICRO & 0.83913 & 0.8258 & 0.82844 & 0.53563 \\ 
    \bottomrule 
    \end{tabular}
\end{table}

\begin{table}[]
    \centering
    \caption{MR, 8000 Training Sequences.}
    \label{tab:my_label}
    \begin{tabular}{lllll}
    \toprule
    & \bfseries WFA & \bfseries LSTM & \bfseries BERT & \bfseries BASELINE \\ 
    \midrule 
    N4 & 0.82277 & 0.81864 & 0.82526 & 0.59726 \\ 
    N5 & 0.84592 & 0.85317 & 0.85399 & 0.55846 \\ 
    MICRO & 0.84395 & 0.85023 & 0.85155 & 0.56176 \\ 
    \bottomrule 
    \end{tabular}
\end{table}

\begin{table}[]
    \centering
    \caption{MSPCP, 400 Training Sequences.}
    \label{tab:my_label}
  \begin{tabular}{lllll}
    \toprule
    & \bfseries WFA & \bfseries LSTM & \bfseries BERT & \bfseries BASELINE \\ 
    \midrule 
    N1 & 0.89774 & 0.81564 & 0.83128 & 0.83008 \\ 
    N2 & 0.83119 & 0.6226 & 0.7294 & 0.55278 \\ 
    N3 & 0.6639 & 0.33845 & 0.51782 & 0.1582 \\ 
    N4 & 0.6367 & 0.23165 & 0.44983 & 0.053211 \\ 
    N5 & 0.71182 & 0.25286 & 0.52513 & 0.0077381 \\ 
    MICRO & 0.68458 & 0.25665 & 0.50207 & 0.040077 \\ 
    \bottomrule 
\end{tabular}
\end{table}

\begin{table}[]
    \centering
    \caption{MSPCP 1000, Training Sequences.}
    \label{tab:my_label}
  \begin{tabular}{lllll}
    \toprule
    & \bfseries WFA & \bfseries LSTM & \bfseries BERT & \bfseries BASELINE \\ 
    \midrule 
    N1 & 0.91579 & 0.90947 & 0.93083 & 0.85113 \\ 
    N2 & 0.84449 & 0.71795 & 0.81302 & 0.67625 \\ 
    N3 & 0.69214 & 0.48918 & 0.61035 & 0.26223 \\ 
    N4 & 0.66321 & 0.38853 & 0.51871 & 0.12028 \\ 
    N5 & 0.76345 & 0.36944 & 0.55099 & 0.042548 \\ 
    MICRO & 0.7258 & 0.38895 & 0.54801 & 0.091939 \\ 
    \bottomrule 
    \end{tabular}
\end{table}

\begin{table}[]
    \centering
    \caption{MSPCP, 2000 Training Sequences.}
    \label{tab:my_label}
  \begin{tabular}{lllll}
    \toprule
    & \bfseries WFA & \bfseries LSTM & \bfseries BERT & \bfseries BASELINE \\ 
    \midrule 
    N1 & 0.94737 & 0.94015 & 0.92541 & 0.7985 \\ 
    N2 & 0.89145 & 0.82017 & 0.86927 & 0.64694 \\ 
    N3 & 0.72948 & 0.57631 & 0.66934 & 0.35078 \\ 
    N4 & 0.69176 & 0.46408 & 0.57831 & 0.18212 \\ 
    N5 & 0.77876 & 0.43644 & 0.60665 & 0.083977 \\ 
    MICRO & 0.74752 & 0.4607 & 0.60521 & 0.14338 \\ 
    \bottomrule 
    \end{tabular}
\end{table}

\begin{table}[]
    \centering
    \caption{MSPCP, 4000 Training Sequences.}
    \label{tab:my_label}
    \begin{tabular}{lllll}
    \toprule
    & \bfseries WFA & \bfseries LSTM & \bfseries BERT & \bfseries BASELINE \\ 
    \midrule 
    N1 & 0.96842 & 0.95368 & 0.94737 & 0.88872 \\ 
    N2 & 0.92282 & 0.83142 & 0.88642 & 0.70724 \\ 
    N3 & 0.77011 & 0.58983 & 0.67558 & 0.46542 \\ 
    N4 & 0.73517 & 0.48084 & 0.56523 & 0.24313 \\ 
    N5 & 0.83144 & 0.47518 & 0.57467 & 0.11361 \\ 
    MICRO & 0.79601 & 0.48997 & 0.58302 & 0.19071 \\ 
    \bottomrule 
    \end{tabular}
\end{table}

\begin{table}[]
    \centering
    \caption{MSPCP, 8000 Training Sequences.}
    \label{tab:my_label}
    \begin{tabular}{lllll}
    \toprule
    & \bfseries WFA & \bfseries LSTM & \bfseries BERT & \bfseries BASELINE \\ 
    \midrule 
    N1 & 0.98195 & 0.97714 & 0.97143 & 0.91729 \\ 
    N2 & 0.93024 & 0.90049 & 0.91239 & 0.73807 \\ 
    N3 & 0.78783 & 0.67781 & 0.70733 & 0.59942 \\ 
    N4 & 0.7477 & 0.53759 & 0.56915 & 0.35323 \\ 
    N5 & 0.8491 & 0.50004 & 0.54256 & 0.1588 \\ 
    MICRO & 0.81193 & 0.53088 & 0.56849 & 0.2643 \\ 
    \bottomrule 
    \end{tabular}
\end{table}

\begin{table}[]
    \centering
    \caption{MSPC-U, 400 Training Sequences.}
    \label{tab:my_label}
    \begin{tabular}{llll}
    \toprule
    & \bfseries WFA & \bfseries LSTM & \bfseries BERT \\ 
    \midrule 
    N2 & 0.77232 & 0.73303 & 0.81822 \\ 
    N3 & 0.63313 & 0.40291 & 0.56004 \\ 
    N4 & 0.34065 & 0.2175 & 0.3016 \\ 
    N5 & 0.14201 & 0.10539 & 0.13042 \\ 
    MICRO & 0.1585 & 0.11474 & 0.14464 \\ 
    \bottomrule 
\end{tabular}
\end{table}

\begin{table}[]
    \centering
    \caption{MSPC-U, 1000 Training Sequences.}
    \label{tab:my_label}
    \begin{tabular}{llll}
    \toprule
    & \bfseries WFA & \bfseries LSTM & \bfseries BERT \\ 
    \midrule 
    N2 & 0.81324 & 0.82896 & 0.83926 \\ 
    N3 & 0.64174 & 0.48128 & 0.59383 \\ 
    N4 & 0.34826 & 0.29307 & 0.32861 \\ 
    N5 & 0.14637 & 0.13186 & 0.13814 \\ 
    MICRO & 0.16311 & 0.14516 & 0.15393 \\ 
    \bottomrule 
    \end{tabular}
\end{table}

\begin{table}[]
    \centering
    \caption{MSPC-U, 2000 Training Sequences.}
    \label{tab:my_label}
    \begin{tabular}{llll}
    \toprule
    & \bfseries WFA & \bfseries LSTM & \bfseries BERT \\ 
    \midrule 
    N2 & 0.8269 & 0.83715 & 0.84263 \\ 
    N3 & 0.65265 & 0.53285 & 0.63791 \\ 
    N4 & 0.35574 & 0.31071 & 0.349 \\ 
    N5 & 0.14912 & 0.13942 & 0.14534 \\ 
    MICRO & 0.16625 & 0.15358 & 0.16222 \\ 
\bottomrule 
\end{tabular}
\end{table}

\begin{table}[]
    \centering
    \caption{MSPC-U, 4000 Training Sequences.}
    \label{tab:my_label}
    \begin{tabular}{llll}
    \toprule
    & \bfseries WFA & \bfseries LSTM & \bfseries BERT \\ 
    \midrule 
    N2 & 0.76113 & 0.71213 & 0.75573 \\ 
    N3 & 0.64381 & 0.53365 & 0.59127 \\ 
    N4 & 0.3578 & 0.32639 & 0.33669 \\ 
    N5 & 0.15151 & 0.14394 & 0.14514 \\ 
    MICRO & 0.1686 & 0.15897 & 0.16098 \\ 
\bottomrule 
\end{tabular}
\end{table}

\begin{table}[]
    \centering
    \caption{MSPC-U, 8000 Training Sequences.}
    \label{tab:my_label}
    \begin{tabular}{llll}
    \toprule
    & \bfseries WFA & \bfseries LSTM & \bfseries BERT \\ 
    \midrule 
    N2 & 0.55257 & 0.45058 & 0.54054 \\ 
    N3 & 0.5661 & 0.49899 & 0.53307 \\ 
    N4 & 0.34073 & 0.32207 & 0.33099 \\ 
    N5 & 0.14853 & 0.14779 & 0.14838 \\ 
    MICRO & 0.16437 & 0.16211 & 0.16341 \\ 
    \bottomrule 
    \end{tabular}
\end{table}

\begin{table}[]
    \centering
    \caption{MR-U, 400 Training Sequences.}
    \label{tab:my_label}
    \begin{tabular}{llll}
    \toprule
& \bfseries WFA & \bfseries LSTM & \bfseries BERT \\ 
\midrule 
N4 & 0.78938 & 0.68689 & 0.75781 \\ 
N5 & 0.81158 & 0.73209 & 0.7868 \\ 
MICRO & 0.80984 & 0.72854 & 0.78452 \\ 
\bottomrule 
\end{tabular}
\end{table}

\begin{table}[]
    \centering
    \caption{MR-U, 1000 Training Sequences.}
    \label{tab:my_label}
    \begin{tabular}{llll}
\toprule
& \bfseries WFA & \bfseries LSTM & \bfseries BERT \\ 
\midrule 
N4 & 0.79892 & 0.753 & 0.78288 \\ 
N5 & 0.82284 & 0.79121 & 0.80471 \\ 
MICRO & 0.82096 & 0.7882 & 0.803 \\ 
\bottomrule 
\end{tabular}
\end{table}

\begin{table}[]
    \centering
    \caption{MR-U, 2000 Training Sequences.}
    \label{tab:my_label}
    \begin{tabular}{llll}
    \toprule
    & \bfseries WFA & \bfseries LSTM & \bfseries BERT \\ 
    \midrule 
    N4 & 0.80935 & 0.77132 & 0.80411 \\ 
    N5 & 0.83072 & 0.80961 & 0.82234 \\ 
    MICRO & 0.82903 & 0.8066 & 0.82091 \\ 
    \bottomrule 
    \end{tabular}
\end{table}

\begin{table}[]
    \centering
    \caption{MR-U, 4000 Training Sequences.}
    \label{tab:my_label}
    \begin{tabular}{llll}
    \toprule
    & \bfseries WFA & \bfseries LSTM & \bfseries BERT \\ 
    \midrule 
    N4 & 0.81769 & 0.79101 & 0.79967 \\ 
    N5 & 0.83928 & 0.82256 & 0.82509 \\ 
    MICRO & 0.83758 & 0.82008 & 0.82309 \\ 
    \bottomrule 
    \end{tabular}
\end{table}

\begin{table}[]
    \centering
    \caption{MR-U, 8000 Training Sequences.}
    \label{tab:my_label}
    \begin{tabular}{llll}
    \toprule
    & \bfseries WFA & \bfseries LSTM & \bfseries BERT \\ 
    \midrule 
    N4 & 0.8205 & 0.80457 & 0.81277 \\ 
    N5 & 0.8425 & 0.84127 & 0.84245 \\ 
    MICRO & 0.84077 & 0.83838 & 0.84012 \\ 
    \bottomrule 
    \end{tabular}
\end{table}

\begin{table}[]
    \centering
    \caption{MSPCP-U, 400 Training Sequences.}
    \label{tab:my_label}
    \begin{tabular}{llll}
    \toprule
    & \bfseries WFA & \bfseries LSTM & \bfseries BERT \\ 
    \midrule 
    N2 & 0.70856 & 0.54879 & 0.62302 \\ 
    N3 & 0.64968 & 0.31778 & 0.49905 \\ 
    N4 & 0.63467 & 0.22759 & 0.44711 \\ 
    N5 & 0.71121 & 0.25239 & 0.52548 \\ 
    MICRO & 0.68721 & 0.24644 & 0.50169 \\ 
    \bottomrule 
    \end{tabular}
\end{table}

\begin{table}[]
    \centering
    \caption{MSPCP-U, 1000 Training Sequences.}
    \label{tab:my_label}
    \begin{tabular}{llll}
    \toprule
    & \bfseries WFA & \bfseries LSTM & \bfseries BERT \\ 
    \midrule 
    N2 & 0.56343 & 0.63362 & 0.61802 \\ 
    N3 & 0.66231 & 0.43854 & 0.56343 \\ 
    N4 & 0.65681 & 0.37763 & 0.5092 \\ 
    N5 & 0.7621 & 0.36656 & 0.54989 \\ 
    MICRO & 0.72875 & 0.37139 & 0.53812 \\ 
    \bottomrule 
    \end{tabular}
\end{table}

\begin{table}[]
    \centering
    \caption{MSPCP-U, 2000 Training Sequences.}
    \label{tab:my_label}
\begin{tabular}{llll}
    \toprule
    & \bfseries WFA & \bfseries LSTM & \bfseries BERT \\ 
    \midrule 
    N2 & 0.52016 & 0.65202 & 0.64873 \\ 
    N3 & 0.67633 & 0.49721 & 0.60039 \\ 
    N4 & 0.67863 & 0.43764 & 0.55913 \\ 
    N5 & 0.77518 & 0.42648 & 0.60156 \\ 
    MICRO & 0.74444 & 0.43131 & 0.58896 \\ 
    \bottomrule 
    \end{tabular}
\end{table}

\begin{table}[]
    \centering
    \caption{MSPCP-U 4000 Training Sequences.}
    \label{tab:my_label}
    \begin{tabular}{llll}
    \toprule
    & \bfseries WFA & \bfseries LSTM & \bfseries BERT \\ 
    \midrule 
    N2 & 0.52822 & 0.58149 & 0.54327 \\ 
    N3 & 0.69234 & 0.45822 & 0.56041 \\ 
    N4 & 0.71764 & 0.44072 & 0.53155 \\ 
    N5 & 0.82694 & 0.46119 & 0.56332 \\ 
    MICRO & 0.79166 & 0.45506 & 0.55384 \\ 
    \bottomrule 
    \end{tabular}
\end{table}

\begin{table}[]
    \centering
    \caption{MSPCP-U, 8000 Training Sequences.}
    \label{tab:my_label}
    \begin{tabular}{llll}
    \toprule
    & \bfseries WFA & \bfseries LSTM & \bfseries BERT \\ 
    \midrule 
    N2 & 0.81084 & 0.59462 & 0.24505 \\ 
    N3 & 0.6507 & 0.46172 & 0.48333 \\ 
    N4 & 0.71584 & 0.4504 & 0.49123 \\ 
    N5 & 0.84148 & 0.46038 & 0.51201 \\ 
    MICRO & 0.80016 & 0.45745 & 0.50523 \\ 
    \bottomrule 
    \end{tabular}
\end{table}

\end{document}